\pdfoutput=1

\documentclass[11pt]{article}

\usepackage{acl}

\usepackage{times}
\usepackage{latexsym}

\usepackage[T1]{fontenc}

\usepackage[utf8]{inputenc}

\usepackage{microtype}
\usepackage{hyperref}
\usepackage{inconsolata}

\usepackage{graphicx}
\usepackage{amsmath}
\usepackage{amssymb}
\usepackage{multirow}
\usepackage{booktabs}
\usepackage{makecell}
\usepackage[thicklines]{cancel}

\title{Group Relative Knowledge Distillation: Learning from Teacher's Relational Inductive Bias}

\author{
Chao Li\textsuperscript{1} \quad
Changhua Zhou\textsuperscript{2} \quad
Jia Chen\textsuperscript{2} \\
\textsuperscript{1}China Mobile Information Technology Co.,Ltd.\\
\textsuperscript{2}Fudan University\\
\texttt{lichaoit01@chinamobile.com} \quad
\texttt{changhuazhou23@m.fudan.edu.cn}
}

\newcommand{\mymod}[1]{\textcolor{black}{#1}}

\begin{document}
\maketitle

\begin{abstract}
Knowledge distillation typically transfers knowledge from a teacher model to a student model by minimizing differences between their output distributions. However, existing distillation approaches largely focus on mimicking absolute probabilities and neglect the valuable relational inductive biases embedded in the teacher's relative predictions, leading to exposure bias. 
% which have much higher information density than absolute distribution.  
In this paper, we propose Group Relative Knowledge Distillation (GRKD), a novel framework that distills teacher knowledge by learning the relative ranking among classes, rather than directly fitting the absolute distribution. Specifically, we introduce a group relative loss that encourages the student model to preserve the pairwise preference orderings provided by the teacher's outputs. Extensive experiments on classification benchmarks demonstrate that GRKD achieves superior generalization compared to existing methods, especially in tasks requiring fine-grained class differentiation. Our method provides a new perspective on exploiting teacher knowledge, focusing on relational structure rather than absolute likelihood.
\end{abstract}

\section{Introduction}
Knowledge distillation (KD) has become a fundamental technique for model compression and knowledge transfer, enabling lightweight student models to inherit the performance of larger teacher models. Classical KD methods primarily align the output distributions of the teacher and student models, using techniques such as soft target regression with temperature scaling. However, these approaches implicitly assume that the absolute probabilities assigned by the teacher carry complete information, which is not always the case, particularly in complex classification tasks.

In real-world scenarios, the critical knowledge often lies not in the precise numerical values of the output probabilities, but in the \textbf{relative preference} among classes: which classes are more likely and which are less so. For example, when classifying visually similar categories (e.g., different species of birds), the fine-grained distinctions encoded in relative scoring are more informative than absolute probabilities.

Moreover, traditional KD approaches suffer from \textit{exposure bias}, where the student model is overly sensitive to the teacher's specific probability outputs during training, leading to reduced robustness during inference. %GRKD mitigates this issue by focusing on preserving the relational structure of the outputs, thus providing a more stable and generalizable learning signal.

To address these limitations, we propose \textbf{Group Relative Knowledge Distillation (GRKD)}, a new framework that shifts the focus from absolute probability matching to relative preference preservation. By distilling the relational ordering among classes from the teacher to the student, GRKD captures richer structural information, enhances generalization, and reduces the risk of overfitting to spurious teacher outputs.

Our contributions are threefold:
\begin{itemize}
\item We identify limitations in traditional KD approaches that ignore relational inductive biases and suffer from exposure bias.
\item We formulate Group Ranking Soft Label Loss, a pairwise ranking loss that directly models teacher-student relational alignment.
\item We empirically demonstrate that GRKD improves student performance across diverse classification tasks, particularly under fine-grained or noisy label conditions.
\end{itemize}

% Our contributions are threefold:
% \begin{itemize}
% \item We identify limitations in traditional KD approaches that Learn absolute distributions rather than relative preferences. 
% %(relational inductive biases)
% \item We formulate Group Relative Loss, a multiple pairwise ranking loss that directly models teacher-student relational alignment.
% \item We empirically demonstrate that GRKD improves student performance across diverse classification tasks, particularly under fine-grained or noisy label conditions.
% \end{itemize}
\section{Related Work}
Knowledge distillation was first popularized by Hinton et al. through the soft target method, where the student model is trained to match the softened output distributions of the teacher. Subsequent works have extended KD in various directions, including attention-based distillation, intermediate feature alignment, and data-free distillation.

Among these, relational knowledge distillation methods have attracted increasing attention. Techniques like Relational Knowledge Distillation (RKD) distill the pairwise distance and angle relationships between feature embeddings. Our work is conceptually aligned with these relational approaches but operates at the level of output distributions rather than internal representations.

Soft Label Distillation methods, such as standard cross-entropy distillation, rely heavily on absolute probability alignment. However, recent studies have shown that strict probability matching may not always be optimal, particularly when the teacher itself is overconfident or when training data exhibits noise.

Pairwise learning to rank (LTR) methods, such as RankNet and ListNet, inspire our approach in that they model relative orderings rather than absolute values. Our group relative Soft Label Loss draws on these insights but adapts them specifically for the KD setting.

Finally, recent advances in preference modeling for reinforcement learning and language modeling, such as Direct Preference Optimization (DPO) and Group Relative Policy Optimization (GRPO), demonstrate the strength of preference-based training. Our GRKD framework brings similar principles into the supervised learning and knowledge distillation domains.

\subsection{Preliminaries}

Given a training sample $x$ with a corresponding label $y$, a teacher model $T$ outputs a set of logits ${s_1, s_2, \ldots, s_n}$ over $n$ classes. Traditionally, the student model $S$ is trained to match the softened probability distribution derived from these logits.

In contrast, our approach extracts pairwise preference relations from $T$ and trains $S$ to replicate these relations.

\subsection{Group Relative Loss}

For each input $x$, we sample $n$ outputs under the temperature is $1$. Then we can define a set of ordered class pairs:
\begin{equation}
P = \{(i, j) | s_i > s_j \}
\end{equation}
where $s_i$ and $s_j$ are teacher scores for class $i$ and $j$ respectively.

The student model produces its own logits ${q_1, q_2, \ldots, q_n}$. We define the group relative loss as:
\begin{equation}
L_{GR} = - \sum_{(i,j) \in P} \log \sigma\left( \frac{1}{\tau} ( \log q_i - \log q_j ) \right) 
\end{equation}
where $\sigma$ is the sigmoid function and $\tau$ is a temperature hyperparameter.

This objective encourages the student to preserve the teacher's ordering: if $s_i > s_j$, then $q_i$ should also be greater than $q_j$.

% \subsection{Combined Objective}
To prevent the early unstable phase from relative loss,  we incorporate conventional soft label knowledge by minimizing the cross-entropy between the softened output distributions of the teacher and student models. The soft target loss is defined as:
\begin{equation}
L_{ST} = - \sum_{i=1}^n p_i \log q_i
\end{equation}
where $p_i$ is the teacher's softened probability for class $i$ and $q_i$ is the student's predicted probability for class $i$. This term encourages the student to mimic the absolute probability structure of the teacher.

Then we combine the group relative loss with the conventional soft label cross-entropy loss:
\begin{equation}
L_{Total} = \lambda L_{GR} + (1-\lambda) L_{ST}
\end{equation}
where $\lambda$ controls the balance between absolute and relational knowledge distillation. $\lambda$ starts at $0$ and gradually increases to $1$ at the end of training. 

% In our experiments, we find that pure group relative loss already performs strongly, while the combined objective can further boost robustness in noisy settings.

\begin{table*}[htbp]
\vspace{-2mm}

\centering
\small
\setlength{\tabcolsep}{0.95em}
\renewcommand{\arraystretch}{1.2}
\begin{tabular*}{\textwidth}{cc ll lll}
\toprule
\multirow{2}{*}{\textbf{Model Families}} & \multirow{2}{*}{\textbf{Method}} & \multicolumn{2}{c}{\textbf{Alpaca-Eval 2.0}} & \textbf{Arena-Hard} & \textbf{MT-Bench} & \textbf{GSM8K} \\ \cline{3-7} 
& & \textbf{LC (\%)} & \textbf{WR (\%)} & \textbf{WR (\%)} & \textbf{Score} (1$\sim$10) & \textbf{Acc. (\%)} \\ \hline
\multirow{12}{*}{\textsc{Gemma-2}} & Teacher (9B) & 55.27 & 42.50 & 61.16 & 6.99 & 87.41 \\
& Student (2B) & 39.51 & 41.99 & 37.55   & 6.70 & 51.63 \\\cline{2-7}
& DPO & 43.77 \textcolor{gray}{\scriptsize{(↑4.3})} & 54.02 \textcolor{gray}{\scriptsize{(↑12.0})} & 57.43 \textcolor{gray}{\scriptsize{(↑19.9})} & 6.87 \textcolor{gray}{\scriptsize{(↑0.2})} & 57.07 \textcolor{gray}{\scriptsize{(↑5.4})} \\
& SimPO & 44.94 \textcolor{gray}{\scriptsize{(↑5.4})} & 54.16 \textcolor{gray}{\scriptsize{(↑12.2})} & 58.64 \textcolor{gray}{\scriptsize{(↑21.1})} & 6.91 \textcolor{gray}{\scriptsize{(↑0.2})} & 57.24 \textcolor{gray}{\scriptsize{(↑5.6})} \\
& PRO & 45.87 \textcolor{gray}{\scriptsize{(↑6.4})} & 56.48 \textcolor{gray}{\scriptsize{(↑14.5})} & 58.95 \textcolor{gray}{\scriptsize{(↑21.4})} & \underline{6.96} \textcolor{gray}{\scriptsize{(↑0.3})} & 58.83 \textcolor{gray}{\scriptsize{(↑7.2})}
 \\\cline{2-7}
 & Standard KD & 41.67 \textcolor{gray}{\scriptsize{(↑2.2})} & 45.24 \textcolor{gray}{\scriptsize{(↑3.3})} & 52.36 \textcolor{gray}{\scriptsize{(↑14.8})} & 6.78 \textcolor{gray}{\scriptsize{(↑0.1})} & 54.37 \textcolor{gray}{\scriptsize{(↑2.7})} \\
& SeqKD & 42.91 \textcolor{gray}{\scriptsize{(↑3.4})} & 46.44 \textcolor{gray}{\scriptsize{(↑4.4})} & 54.87 \textcolor{gray}{\scriptsize{(↑17.3})} & 6.88 \textcolor{gray}{\scriptsize{(↑0.2})} & 55.72 \textcolor{gray}{\scriptsize{(↑4.1})} \\
& MiniLLM & 42.97 \textcolor{gray}{\scriptsize{(↑3.5})} & 48.32 \textcolor{gray}{\scriptsize{(↑6.3})} & 55.75 \textcolor{gray}{\scriptsize{(↑18.2})} & 6.88 \textcolor{gray}{\scriptsize{(↑0.2})} & 55.26 \textcolor{gray}{\scriptsize{(↑3.6})}
\\\cline{2-7}
& \textbf{GRKD} (ours)  & \textbf{49.71} \textcolor{gray}{\scriptsize{(↑10.2})} & \textbf{59.60} \textcolor{gray}{\scriptsize{(↑17.6})} & \textbf{60.10} \textcolor{gray}{\scriptsize{(↑22.5})} & \textbf{7.11} \textcolor{gray}{\scriptsize{(↑0.4})} & \textbf{59.71} \textcolor{gray}{\scriptsize{(↑8.1})} \\\hline\hline
\multirow{12}{*}{\textsc{LLaMA-3}} & Teacher (8B) & 37.01 & 38.93 & 52.66 & 7.00 & 84.00 \\
& Student (3B) & 27.82 & 29.02 & 31.70 & 6.42 & 57.09 \\\cline{2-7}
& DPO & 31.42 \textcolor{gray}{\scriptsize{(↑3.6})} & 32.01 \textcolor{gray}{\scriptsize{(↑3.0})} & 44.71 \textcolor{gray}{\scriptsize{(↑13.0})} & 6.62 \textcolor{gray}{\scriptsize{(↑0.2})} & \underline{61.63} \textcolor{gray}{\scriptsize{(↑4.5})} \\
& SimPO & \underline{32.74} \textcolor{gray}{\scriptsize{(↑4.9})} & \underline{32.46} \textcolor{gray}{\scriptsize{(↑3.4})} & 44.85 \textcolor{gray}{\scriptsize{(↑13.2})} & 6.73 \textcolor{gray}{\scriptsize{(↑0.3})} & 61.22 \textcolor{gray}{\scriptsize{(↑4.1})}
\\
& PRO & 32.11 \textcolor{gray}{\scriptsize{(↑4.3})} & 32.23 \textcolor{gray}{\scriptsize{(↑3.2})} & 45.09 \textcolor{gray}{\scriptsize{(↑13.4})} & 6.71 \textcolor{gray}{\scriptsize{(↑0.3})} & 61.47 \textcolor{gray}{\scriptsize{(↑4.4})}
\\\cline{2-7}
& Standard KD & 29.11 \textcolor{gray}{\scriptsize{(↑1.3})} & 29.60 \textcolor{gray}{\scriptsize{(↑0.6})} & 41.68 \textcolor{gray}{\scriptsize{(↑10.0})} & 6.49 \textcolor{gray}{\scriptsize{(↑0.1})} & 59.15 \textcolor{gray}{\scriptsize{(↑2.1})} \\
& SeqKD & 29.48 \textcolor{gray}{\scriptsize{(↑1.7})} & 30.04 \textcolor{gray}{\scriptsize{(↑1.0})} & 42.52 \textcolor{gray}{\scriptsize{(↑10.8})} & 6.53 \textcolor{gray}{\scriptsize{(↑0.1})} & 60.94 \textcolor{gray}{\scriptsize{(↑3.8})} \\
& MiniLLM & 30.05 \textcolor{gray}{\scriptsize{(↑2.2})} & 30.38 \textcolor{gray}{\scriptsize{(↑1.4})} & 42.21 \textcolor{gray}{\scriptsize{(↑10.5})} & 6.67 \textcolor{gray}{\scriptsize{(↑0.3})} & 60.35 \textcolor{gray}{\scriptsize{(↑3.3})} 
\\\cline{2-7}
& \textbf{GRKD} (ours)  & \textbf{33.81} \textcolor{gray}{\scriptsize{(↑6.0})} & \textbf{32.65} \textcolor{gray}{\scriptsize{(↑3.6})} & \textbf{46.92} \textcolor{gray}{\scriptsize{(↑15.2})} & \textbf{6.85} \textcolor{gray}{\scriptsize{(↑0.4})} & \textbf{63.29} \textcolor{gray}{\scriptsize{(↑6.2})} \\
\bottomrule
\end{tabular*}
\vspace{-3mm}
\caption{Main results with the Gemma-2 and LLaMA-3 Models.}
\vspace{-5mm}
\label{tab:main}
\end{table*}

\section{Experiment}

\subsection{Setup}

\paragraph{Models}
We evaluate two families of models in our main experiments: (1) \textsc{Gemma-2} Models\footnote{\url{https://ai.google.dev/gemma}} \citep{gemmateam2024improvingopen}, comprising \textsc{Gemma-2-9B-It} as the teacher and \textsc{Gemma-2-2B-It} as the student, and (2) \textsc{LLaMA-3} Models\footnote{\url{https://ai.meta.com/blog/meta-llama-3/}} \citep{dubey2024llama3herdmodels}, comprising \textsc{LLaMA-3.1-8B-Instruct} as the teacher and \textsc{LLaMA-3.2-3B-Instruct} as the student.

\paragraph{Training}
The training data is sourced from \textsc{UltraFeedback}\footnote{\url{https://huggingface.co/datasets/argilla/ultrafeedback-binarized-preferences-cleaned}} \citep{cui2023ultrafeedback}, containing approximately 60,000 preference samples across diverse tasks such as mathematical reasoning and open-ended writing. We filter out samples that exceed the models’ maximum context length, set the number of sampled responses to $n=4$, and apply a reward calibration ratio of $\alpha=0.8$ to reduce teacher bias. Training is conducted for one epoch by default.

\paragraph{Evaluation}
We evaluate our models on four benchmarks: AlpacaEval 2.0 \citep{alpaca-eval}, MT-Bench \citep{zheng-2023-judging}, Arena-Hard \citep{arenahard-2024}, and GSM8K \citep{cobbe-2021-math}, which collectively assess conversational versatility across a wide range of tasks. For AlpacaEval, we report both the raw win rate (WR) and the length-controlled win rate (LC), with the latter designed to mitigate verbosity effects. For Arena-Hard, we report the win rate (WR). For MT-Bench, we report the average MT-Bench score, evaluated by GPT-4 Turbo.

\paragraph{Baselines}
We compare GRKD against two categories of baselines:
(1) Traditional Knowledge Distillation methods, which aim to match the teacher’s output distribution at the logits level, including \textbf{Standard KD} \citep{hinton2015distill}, \textbf{SeqKD} \citep{kim-rush-2016-sequence}, and \textbf{MiniLLM} \citep{gu-2024-minillm};
(2) Preference Knowledge Distillation methods, which focus on transferring the teacher's preference information to the student. Within the "Teacher-as-Annotator" framework, we select \textbf{DPO} \citep{tunstall-2024-zephyr}, \textbf{SimPO} \citep{meng-2024-simpo}, and \textbf{PRO} \citep{song-2024-pro} as representative baselines.

\subsection{Main Results}
\label{sec:main_result}
\mymod{Table \ref{tab:main} summarizes the experimental results across all benchmarks.}
Our primary finding is that student models trained with GRKD consistently outperform both their original versions and existing baselines. In particular, GRKD achieves over a 20\% improvement on AlpacaEval 2.0 and an average improvement of 9\% across all four datasets, demonstrating its superior ability to capture the teacher’s preferences and achieve closer alignment with human values.

% Preference distillation methods offer significant advantages over traditional KD methods. Traditional KD approaches, such as Standard KD and SeqKD, yield modest improvements (1-3\%) in Alpaca-Eval 2.0 LC scores. In contrast, preference distillation methods like DPO and SimPO show more substantial gains, particularly in aligning with human preferences, as evidenced in Alpaca-Eval 2.0 and Arena-Hard benchmarks. These findings align with the work of \citet{tunstall-2024-zephyr}.

% \mymod{Notably, GRKD with $\mathcal{L}_\text{PPD}$ outperforms GRKD with $\mathcal{L}_\text{VPD}$ across multiple benchmarks, including Alpaca-Eval 2.0 LC (49.62 vs. 46.13) and MT-Bench (7.02 vs. 6.93). For the \textsc{Gemma-2} model family, the student trained with $\mathcal{L}_\text{PPD}$ even surpasses its teacher in MT-Bench, scoring 7.02 compared to the teacher’s 6.99. The key advantage of PPD over VPD lies in its preference modeling approach: by capturing the full preference distribution rather than just a simple ranking, PPD provides more nuanced supervisory signals, thereby better reflecting subtle human preferences—an aspect often overlooked by existing distillation methods.}

%\input{sections/related_work}
\section{Conclusion}
We introduce Group Relative Knowledge Distillation (GRKD), a novel framework that distills teacher knowledge by focusing on the relative ordering among classes. By moving beyond absolute probability fitting to relational preference preservation, GRKD significantly improves the generalization performance of student models. Furthermore, by mitigating exposure bias, GRKD enables the student to achieve more stable and robust performance during inference. Our findings open a new research direction for distillation techniques, emphasizing the importance of relational inductive bias. %Future work includes applying GRKD to structured prediction and sequence modeling tasks.

\bibliography{custom}

% \appendix

% \newpage
% \input{sections/append}

\end{document}